\documentclass{article} 
\usepackage[final]{colm2026_conference}

\usepackage{microtype}
\usepackage{hyperref}
\usepackage{url}
\usepackage{booktabs}
\usepackage{graphicx}   
\usepackage{amsmath}    
\usepackage{amssymb}    
\usepackage{multirow}   
\usepackage{cleveref}   
\usepackage{xspace}
\usepackage{algorithm}
\usepackage{tikz}
\usepackage{pgfplots}
\pgfplotsset{compat=1.18}
\usetikzlibrary{arrows.meta, shapes.geometric, shapes.misc, fit, positioning, calc, backgrounds, decorations.pathreplacing}

\definecolor{darkblue}{rgb}{0, 0, 0.5}
\hypersetup{colorlinks=true, citecolor=darkblue, linkcolor=darkblue, urlcolor=darkblue}

\title{Function Over Form: Distributional Orthogonalization in Mixture-of-Experts with Replica Expert Mechanism}

\author{Jinfan He$^{1,2}$, Yunzhuo Liu$^{2,\ast}$, Kai Zhang$^{2}$, Weidong Han$^{2}$, Key$^{2}$, Rayying$^{2}$ \\
$^{1}$School of EECS, Peking University, China \\
$^{2}$Tencent Hunyuan, China \\
$\ast$ Corresponding author.}

\begin{document}

\maketitle

\begin{abstract}
    The scaling of LLMs increasingly relies on MoE architectures to decouple active computation from total parameter count. However, the efficacy of MoE is often constrained by expert collapse and representation redundancy, both leading to underutilization of model capacity. To address these challenges, this paper proposes Distributional Orthogonalization Loss (DO-loss), an auxiliary regularization that shifts the focus from static weight diversity to dynamic routing behavior. By representing each expert's token assignment history as a high-dimensional binary load signature, DO-loss penalizes signature overlap to prevent expert collapse while encouraging functional specialization. To align this algorithmic design with system efficiency, we further introduce the Replica Expert Mechanism (REM), which improves load balancing through a two-tiered strategy: adjusting replica expert placement at the global-batch level and performing real-time token dispatching at the micro-batch level. Empirical evaluations demonstrate that our method outperforms the evaluated routing algorithms on downstream tasks for both 4.8BA0.5B and 30BA3B MoE models, while maintaining comparable training efficiency.
\end{abstract}

\section{Introduction}
\label{sec:intro}

Mixture-of-Experts (MoE) architectures have emerged as a standard design for scaling large language models \cite{jiang2024mixtral, liu2024deepseek, yang2025qwen3}. By routing each token to a sparse subset of expert feed-forward networks (FFNs) instead of activating the entire parameter space, these models maintain constant per-token computational costs. This efficiency enables the training and deployment of models with trillions of parameters.

The effectiveness of such sparse activation depends on a sufficient degree of expert specialization, as routing tokens to specific experts yields better performance than averaging the processing results across different experts \cite{shazeer2017outrageously, lepikhin2020gshard}. However, this specialization is often impeded by routing biases that emerge early in training. As popular experts receive more tokens, they update more frequently, creating a feedback loop where a few experts dominate while others remain underutilized. 
Without explicit mechanisms for encouraging inter-expert diversity, standard gradient descent often fails to differentiate expert functions, resulting in redundant expert representations that cover similar domains.

Current approaches tend to address these problems in isolation. Load-balancing losses \cite{fedus2022switch, wang2024auxiliary} mitigate expert collapse by penalizing uneven token distributions. However, by encouraging uniform allocation, they push experts to process tokens irrespective of representational compatibility, thereby suppressing specialization. On the other hand, orthogonality-based approaches \cite{ernie2025technicalreport, lv2025coupling, Liu2024OMoE} attempt to promote functional diversity by enforcing orthogonality among experts in the parameter space. However, these methods do not guarantee that all experts receive input tokens. 
Experts can remain mutually orthogonal while pointing toward regions outside the input manifold. 
Directly combining these techniques often produces conflicting gradient directions that destabilize training~\cite{li2023diversifying}. To prevent conflicts with load-balancing methods, recent approaches~\cite{Hendawy2024MOORE, wang2025advancing} introduce orthogonality constraints within the expert output spaces. However, this incurs significant computational, communication, and memory overhead, and it integrates poorly with standard distributed training frameworks~\cite{rajbhandari2022deepspeed, Shoeybi2019Megatron}.

\begin{figure}[htbp]
    \centering
    \includegraphics[width=0.9\linewidth]{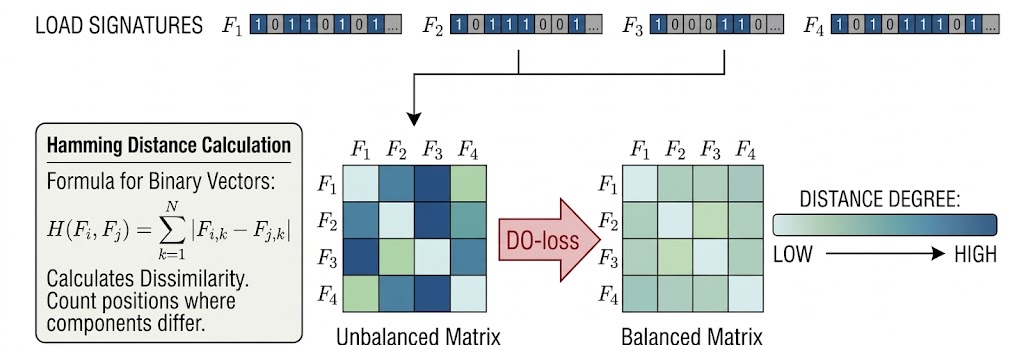}
    \caption{Token assignment histories encoded as binary load signatures. While the total sum of the pairwise Hamming distances remains constant under top-$K$ routing, the DO-loss drives the distances between all pairs of experts toward equidistance to penalize high routing overlap.}
    \label{figs:do_balancing}
\end{figure}

In this paper, we propose the Distributional Orthogonalization loss (DO-loss). The core idea is to record each expert's token assignment history as a binary load signature and penalize small pairwise Hamming distances between signatures, which indicate routing overlap (Fig.~\ref{figs:do_balancing}). This directly encourages specialization: experts covering similar token domains develop highly overlapping signatures and thus incur severe penalties, forcing them to diverge. The same mechanism also makes expert collapse self-correcting. When multiple experts become idle, the fixed total pairwise distance (Eq.~\ref{eq:zero_sum}) forces larger distances among the active experts, which increases distance variance. Minimizing this variance under the zero-sum constraint pushes the router to re-activate idle experts. Through a single loss term operating on routing statistics rather than model weights, DO-loss thus addresses both collapse and specialization with negligible computational overhead.

While DO-loss prevents expert collapse, it does not guarantee strict load balancing. Because specialized experts attract different volumes of tokens, load imbalance is an inherent byproduct of specialization. In expert-parallel deployments, this imbalance has two interacting effects: overloaded ranks become stragglers that limit overall throughput, while the excess activations they must hold create memory footprint disparities that increase the risk of out-of-memory (OOM) failures at scale. To close this gap, we introduce the Replica Expert Mechanism (REM). REM first replicates hot experts onto underloaded ranks to rebalance the static workload, and then routes incoming tokens to the shortest available queue at runtime to absorb residual fluctuations. This allows the system to maintain near-uniform resource utilization without constraining expert specialization.

We evaluate our approach on 4.8BA0.5B and 30BA3B MoE models across six downstream benchmarks. By synchronizing the DO-loss distance matrix across data-parallel ranks, Global-DO-loss operates at the global-batch level and outperforms all baselines at both scales. Our method simultaneously resolves expert collapse and expert representation redundancy. REM reduces load imbalance below the level attained by traditional load balancing methods, recovering the throughput lost to specialization without any degradation in model quality.

\section{Related Work}

\subsection{Routing and Auxiliary Load Balancing Loss}
Token-choice routing~\cite{fedus2022switch} is standard in autoregressive LMs, where each token independently selects its Top-$K$ experts, but is prone to collapse without constraints. The Switch Transformer~\cite{fedus2022switch} penalizes the product of per-expert routing fraction and mean probability; ST-MoE~\cite{zoph2022st} adds a router z-loss for stability. Qiu et al.~\cite{qiu2025demons} show that computing these losses per device forces local uniformity that hinders specialization, and propose global synchronization to preserve semantic clustering. DeepSeek~\cite{wang2024auxiliary} uses per-expert bias terms updated by load deviation for implicit, gradient-free balancing. Expert-choice routing~\cite{zhou2022mixture} guarantees perfect balance but introduces causal leakage in autoregressive settings~\cite{muennighoff2409olmoe}.

\subsection{Orthogonality in MoE}
Orthogonality-based methods reduce representation redundancy by penalizing similarity among expert parameters or outputs. MOORE~\cite{Hendawy2024MOORE} uses Gram-Schmidt regularization; AES-MoE~\cite{wang2025advancing} regularizes output similarity without affecting load balance. However, the associated overhead restricts these methods to post-training or fine-tuning. ERNIE~\cite{ernie2025technicalreport} and ERC-loss~\cite{lv2025coupling} reduce this cost by operating on routing representations or coupling expert and router optimization.

\subsection{System Efficiency and Expert Scheduling}
Skewed token distributions require robust system-level scheduling. EfficientMoE~\cite{zeng2025efficientmoe} places replicas at the global-batch level but lacks per-step granularity. Pro-Prophet~\cite{wang2026pro} enables real-time micro-batch re-dispatch, though its fine-grained redistribution is time-consuming and its overlap of scheduling with computation depends on the compute-to-communication ratio. 

\section{Methods}
\label{sec:method}

\subsection{Distributional Orthogonalization Loss (DO-loss)}

\subsubsection{Binary Load and Zero-Sum Constraint}
For a batch of $m$ tokens, the workload of expert $i$ is a binary vector $F_i \in \{0,1\}^m$:
\begin{equation}
f_{it} = \mathbb{I}\left(i \in \text{Top-}K(\mathbf{x}_t)\right), \quad t \in \{1, \ldots, m\}
\end{equation}
Let $s_{it}$ denote the raw Sigmoid-gated routing score for expert $i$ on token $t$. The normalized routing probability is $p_{it} = s_{it} / \sum_{i'} s_{i't}$, which satisfies $\sum_i p_{it} = 1$. After Top-$K$ selection, only the $p_{it}$ of the $K$ chosen experts retain their normalized values; all others are set to zero and the $K$ values are re-normalized.

With Top-$K$ routing and $n$ experts, exactly $k$ bits are set per token, so the total pairwise Hamming distance over the batch is fixed:
\begin{equation}
\label{eq:zero_sum}
\sum_{i=1}^n \sum_{j=1}^n \| F_i - F_j \|_1 = 2mk(n-k)
\end{equation}
This zero-sum constraint is the structural key: minimizing the \emph{variance} of pairwise distances drives all pairs toward the same distance without requiring an explicit target. It also explains DO-loss's anti-collapse property. When some experts are idle ($F_i = \mathbf{0}$), the zero-sum constraint forces the active experts to exhibit larger pairwise distances to maintain the fixed total, which increases this variance. Since DO-loss minimizes this variance, idle-expert configurations are inherently penalized: the gradient of the loss with respect to the routing probability $p_{it}$ of an idle expert $i$ is negative for tokens assigned to active experts, pushing the router to re-activate the idle expert. A formal derivation is provided in Appendix~\ref{appendix:gradient_analysis}. The original loss is:
\begin{equation}
\label{eq:origin_do}
\mathcal{L}_{\text{origin\_DO}} = \sum_{i \neq j} \| F_i - F_j \|_1^2
\end{equation}

\subsubsection{Equalization Loss and Gradient Derivation}
Discrete assignments block gradient flow. We apply Straight-Through Estimation (STE) by replacing each $F_i$ with $P_i + \text{sg}[F_i - P_i]$, where $P_i$ is the continuous routing probability vector and $\text{sg}[\cdot]$ denotes the stop-gradient operator, which yields a zero gradient. Substituting into Eq.~\ref{eq:origin_do}:
\begin{equation}
\mathcal{L}_{\text{DO-STE}} = \sum_{i \neq j} \left\| P_i + \text{sg}\!\left[ F_i - P_i \right] -  P_j - \text{sg}\!\left[F_j - P_j\right] \right\|_1^2
\end{equation}
Taking the gradient and noting that $f_{it} \in \{0,1\}$ implies $\text{sgn}(f_{it} - f_{jt}) = f_{it} - f_{jt}$:
\begin{equation}
\nabla \mathcal{L}_{\text{DO-STE}} = \sum_{i \neq j} 2 \| F_i - F_j \|_1 \sum_{t=1}^m (f_{it} - f_{jt}) \nabla (p_{it} - p_{jt})
\end{equation}

\subsubsection{Surrogate Loss Implementation}
We factor the loss into a detached magnitude $G_{ij}$ and a gradient direction term $C_{ij}$:
\begin{equation}
\label{eq:gij_cij}
G_{ij} = \| F_i - F_j \|_1, \quad C_{ij} = \sum_{t=1}^m p_{it}(f_{it} - f_{jt})
\end{equation}
Since $C_{ii} = 0$, the $i \neq j$ constraint is automatically satisfied. With a normalization scalar $\frac{1}{m^2k}$ (derived in Appendix~\ref{appendix:do_normalization}), the final loss is:
\begin{equation}
\label{eq:do_loss}
\mathcal{L}_{\text{DO}} = \frac{1}{m^2k} \sum_{i=1}^n \sum_{j=1}^n G_{ij} \cdot C_{ij}
\end{equation}

\subsubsection{Global-Level Regularization via Cross-DP Synchronization}
\citet{qiu2025demons} show that routing statistics over larger batches produce more stable load distributions. DO-loss achieves the same effect by exploiting its decoupled structure (Eq.~\ref{eq:do_loss}). We synchronize the distance matrix $G_{ij}$ across the DP group via All-Reduce and keep the gradient direction indicator $C_{ij}$ local:
\begin{equation}
G_{ij}^{d} = \| F_i^{d} - F_j^{d} \|_1, \qquad
G_{ij}^{g} = \frac{1}{|\mathcal{D}|} \sum_{d \in \mathcal{D}} G_{ij}^{d}
\end{equation}
The global DO-loss on device $d$ then applies the shared magnitude to the local direction indicator:
\begin{equation}
\label{eq:global_do}
\mathcal{L}_{\text{global\_DO}}^{(d)} = \frac{1}{m^2k} \sum_{i=1}^n \sum_{j=1}^n G_{ij}^{g} \cdot C_{ij}^{d}
\end{equation}

The loss computation complexity of $\mathcal{O}(mn^2)$ is practically negligible because the expert count $n$ (64--256) is far smaller than the hidden dimension, and the quadratic dependence is on $n$ rather than the model width.

The overall training objective is:
\begin{equation}
\mathcal{L}_{total} = \mathcal{L}_{CE} + \lambda \cdot \mathcal{L}_{\text{DO}}
\end{equation}
where $\lambda$ is the DO-loss coefficient and $\mathcal{L}_{CE}$ is the standard cross-entropy loss. No additional load-balancing auxiliary terms are used, as DO-loss alone addresses both collapse and specialization.

\subsection{Replica Expert Mechanism (REM)}
MoE training exhibits stable phases during which hot-expert identities remain consistent across steps \cite{cong2024prediction}. REM uses this property by replicating hot experts to distribute their token load and subsequently routing tokens to the least-loaded replicas in real time.

\begin{figure}[htbp]
  \centering
  \includegraphics[width=0.88\linewidth]{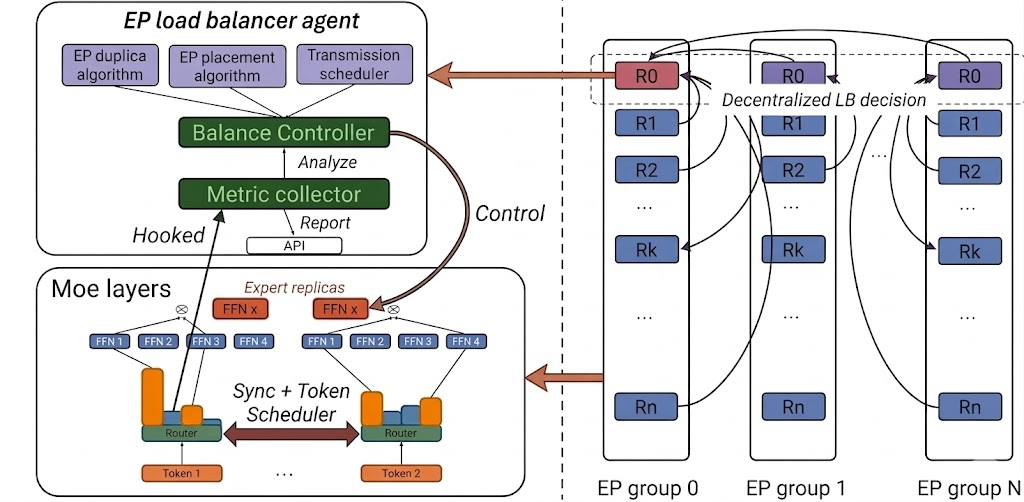}
  \caption{The architecture of REM. \textbf{Left:} The global-batch controller (top) allocates replicas once per global batch; a collector asynchronously retrieves metrics from the router. The micro-batch dispatcher (bottom) synchronizes router information and routes tokens in real time. \textbf{Right:} Load statistics are gathered hierarchically---the coordinator (R0) of each EP group aggregates intra-group rank loads locally, and R0 nodes subsequently exchange summaries across EP groups to construct a global view. This two-level aggregation avoids all-to-all collection across all ranks while enabling decentralized load-balancing decisions at every rank in real time.}
  \label{fig:rem}
\end{figure}

This mechanism operates through two decoupled loops (pseudocode in Appendix~\ref{appendix:rem_pseudocode}; architecture in Figure~\ref{fig:rem}). A metric collector asynchronously aggregates EP load statistics using a moving average window. At each global-batch boundary, a controller evaluates these statistics to greedily place replicas on underloaded ranks based on a Benefit Score (Eq.~\ref{eq:benefit_score}). Within each micro-batch, a local dispatcher routes tokens independently of the global controller to prevent synchronization bottlenecks.

\subsubsection{Global-Batch Replica Allocation}

REM calculates a Benefit Score $S$ for replica placement. The score $S$ represents the optimal number of tokens that the receiving replica $r_r$ can accept from $r_s$ to maximize load balance with replica expert $e$.

Assuming a capacity of $D$ replica expert slots per rank, the scheduler iteratively identifies the least-loaded rank as the receiver $r_r$ and greedily assigns the expert that maximizes $S$. This allocation process continues until all available slots across every rank are fully occupied.

We define the ideal token threshold $\tau$ as the average load across all ranks $\tau = \frac{1}{|\mathcal{R}|} \sum_{r \in \mathcal{R}} L_r $. The score $S$ is then calculated as:
\begin{equation}
\label{eq:benefit_score}
S(e, r_s, r_r, \tau) =
\begin{cases}
\min(\tau - L_{r}, T_e, L_{s} - \tau), & \text{if } L_{r} < \tau \text{ and } L_{s} > \tau \\
\min\left(\left\lfloor \frac{L_{s} - L_{r}}{2} \right\rfloor, T_e\right), & \text{if } L_{r} \ge \tau \text{ and } L_{s} > L_{r} \\
-1, & \text{otherwise}
\end{cases}
\end{equation}
where $L_s$ and $L_r$ denote the historical loads of the sending and receiving ranks, respectively, while $T_e$ represents the token volume of expert $e$. The scheduler identifies the $(e, r_s)$ pair that maximizes $S$, updating the load estimates before the subsequent iteration.

\subsubsection{Dynamic Routing via Micro-step Scheduling}

Although token scheduling at the micro-batch level adheres to the same greedy principle as global-batch allocation, it evaluates the real-time load distribution instead of historical statistics. To obtain a globally consistent instantaneous view, all GPUs synchronize the current depths of the queues via an AllGather operation at the beginning of each micro-batch. This fine-grained signal enables the dispatcher to route tokens dynamically in response to transient load fluctuations that a sliding-window average inherently misses. The overhead of this AllGather operation is negligible compared to the AlltoAll dispatch operation. 

The dispatcher executes exactly $D$ iterations. In each iteration, it sorts the ranks according to the current queue depths and processes them in order from the lightest load to the heaviest. During this traversal, tokens are greedily routed to the expert replica that yields the maximum score for each rank. To avoid prohibitive on-device overhead, the dispatcher does not resort the ranks after each individual allocation. The entire sequence of token dispatch executes on the device without requiring intervention from the global controller.

REM operates purely as a system-level scheduler and does not alter the mathematical training trajectory. Replicas hold asynchronously synchronized copies of the original expert parameters; gradients computed on replicas are accumulated back to the original rank after each global step, with optimizer states residing exclusively on the original rank. A 10,000-step precision test confirmed gradient alignment to three decimal places. Appendix~\ref{appendix:rem_pseudocode} provides the complete pseudocode and formal equivalence conditions.

\begin{figure}[htbp]
    \centering
    \includegraphics[width=0.75\linewidth]{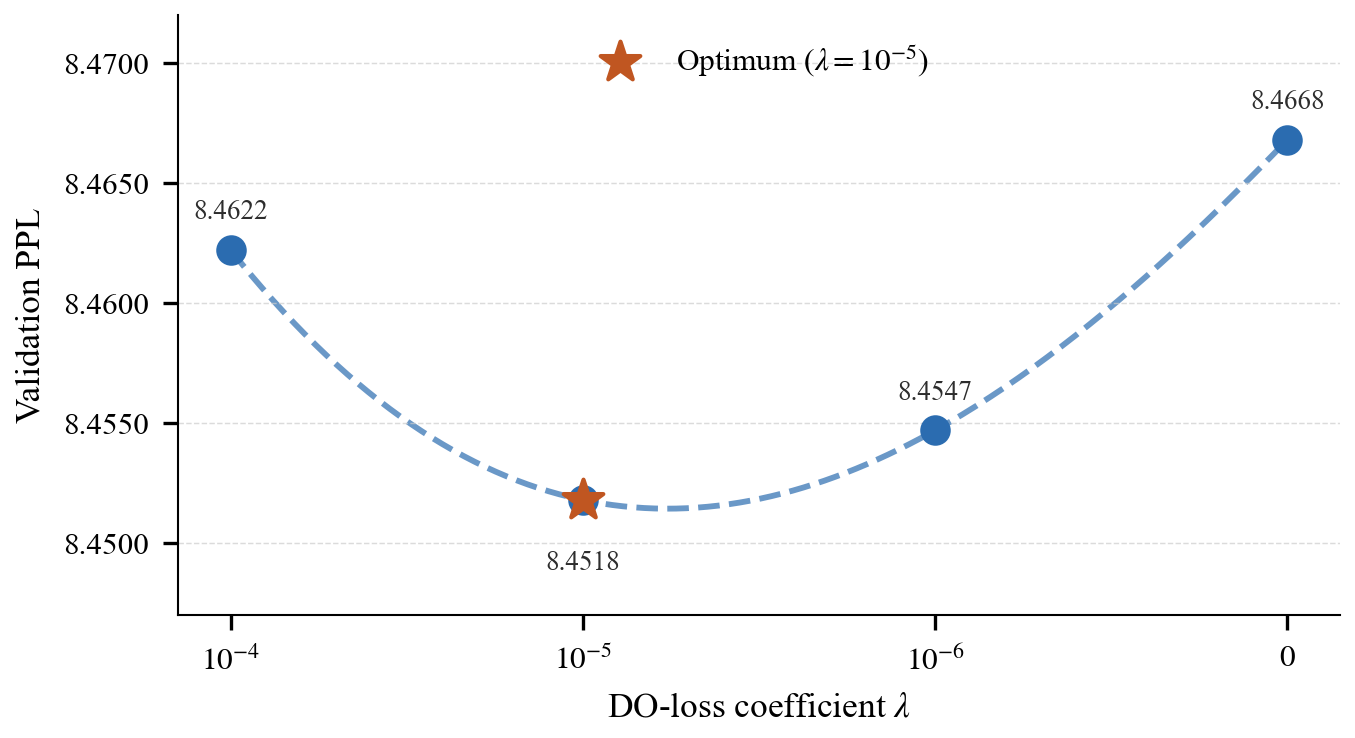}
    \caption{\textbf{Ablation of DO-loss coefficient $\lambda$.} Validation perplexity as a function of $\lambda$ (4.8BA0.5B, 100B tokens). $\lambda{=}0$ disables DO-loss, leading to expert collapse via redundancy. Excessively large $\lambda$ causes the high-dimensional pairwise regularization to overwhelm the cross-entropy signal. The optimum $\lambda{=}10^{-5}$ is approximately two orders of magnitude smaller than typical load-balancing loss coefficients.}
    \label{fig:do_balancing}
\end{figure}

\section{Experimental Setup}
\label{sec:setup}

\subsection{Model Architecture}
We use the Qwen3MoE architecture~\cite{yang2025qwen3} at two scales, following its Active Parameter (A) naming convention. Both models use Sigmoid-gated Top-8 routing over 128 experts ($K{=}8$, $N{=}128$), a vocabulary of 120,832, and 32-head GQA with 4 query groups. The 4.8BA0.5B model has 15 layers and a hidden size of 1024; the 30BA3B model has 48 layers and a hidden size of 2048. Expert intermediate dimension is 768 for both.

\subsection{Baseline Methods}

We compare against five routing strategies spanning the main paradigms.

\textbf{Local-LBL} uses the Switch Transformer auxiliary loss~\cite{fedus2022switch} $\mathcal{L}_{\text{LBL}} = \sum_i f_i p_i$, where $f_i$ is the per-expert routing fraction and $p_i$ the mean routing probability, computed independently on each device.
\textbf{Global-LBL}~\cite{qiu2025demons} synchronizes $f_i$ across data-parallel ranks via All-Reduce before computing the same loss, preserving semantic clustering at the macro level while enforcing global balance.
\textbf{Auxfree}~\cite{wang2024auxiliary} maintains per-expert bias terms $b_i$ updated proportionally to the deviation from the ideal uniform load $b_i \leftarrow b_i + \eta \cdot (L_i - \bar{L})$, achieving balance through implicit feedback without gradient interference.
\textbf{Orth-loss}~\cite{ernie2025technicalreport} penalizes the cosine similarity between normalized routing weight vectors: $\mathcal{L}_{\text{orth}} = \sum_{i \neq j} (\hat{w}_i^\top \hat{w}_j)^2$.
\textbf{Seq-LBL} computes $\mathcal{L}_{\text{LBL}}$ at the per-sequence level with coefficient $10^{-3}$, included only in MaxVio and throughput evaluations as an extreme balancing baseline.

\subsection{Training Settings}
All experiments use the Megatron-Core framework on NVIDIA H20 GPUs with SwiGLU expert FFNs~\cite{shazeer2020glu}, AdamW optimizer ($\beta_1{=}0.9$, $\beta_2{=}0.95$, weight decay 0.1), and a cosine-decay learning rate schedule peaking at $3{\times}10^{-4}$ and decaying to $3{\times}10^{-5}$ after 2000 linear warmup steps. Sequence length is 4096, global batch size 512 (micro-batch 4). The 4.8BA0.5B and 30BA3B models train for 100B and 200B tokens, respectively.

Models are trained on a 2T-token proprietary corpus. Table~\ref{tab:dataset} summarizes its domain composition.

\begin{table}[htbp]
\centering
\small
\begin{tabular}{@{}l l@{}}
\toprule
Domain & Description \\
\midrule
GitHub code & Multi-language code repositories \\
STEM data & Academic papers and textbooks \\
Chinese/English web & General web crawl \\
Academic books & Multi-discipline textbooks \\
Encyclopedias & Encyclopedia-style knowledge \\
Q\&A banks & Exam and quiz datasets \\
\bottomrule
\end{tabular}
\caption{Domain composition of the pretraining corpus.}
\label{tab:dataset}
\end{table}

\subsection{Evaluation Benchmarks}
We evaluate on six benchmarks spanning reasoning, knowledge, and mathematics. AGIEval~\cite{zhong2024agieval} tests human-level reasoning with problems from standardized exams. MMLU~\cite{Hendrycks2021MMLU} and MMLU-Pro~\cite{wang2024mmlu} measure broad multitask language understanding, with the latter offering a more challenging and discriminative variant. CEval~\cite{huang2023c} covers Chinese-language academic knowledge across multiple disciplines. GSM8K~\cite{Cobbe2021GSM8K} assesses multi-step arithmetic reasoning via grade-school math problems. TriviaQA~\cite{Joshi2017TriviaQA} evaluates open-domain factual recall from trivia questions.

\begin{table}[t]
\centering
\small
\setlength{\tabcolsep}{3pt}
\renewcommand{\arraystretch}{1.1}
\begin{tabular}{@{}l l c c c c c c c@{}}
\toprule
Model & Method & AGIEval & MMLU & MMLU-PRO & CEval & GSM8K & TriviaQA & PPL ($\downarrow$) \\
\midrule
\multirow{6}{*}{4.8BA0.5B}
& Local-LBL & 22.24 & 36.01 & 10.95 & 36.63 & 12.71 & 23.56 & 8.4650 \\
& Global-LBL & 23.02 & 36.80 & 11.96 & 37.24 & 13.19 & 23.75 & 8.4604 \\
& Orth-loss & 22.60 & 34.44 & 11.42 & 34.63 & 12.25 & 23.47 & 8.4664 \\
& Auxfree & 23.55 & 36.85 & 12.02 & 38.63 & 13.57 & 23.06 & 8.4620 \\
& DO-loss & 24.05 & 36.33 & 13.50 & 37.10 & 13.51 & 23.89 & 8.4518 \\
& Global-DO-loss & 24.13 & 37.18 & 13.88 & 39.64 & 14.55 & 24.03 & 8.4433 \\
\midrule
\multirow{3}{*}{30BA3B}
& Global-LBL & 41.06 & 54.23 & 21.05 & 57.32 & 56.07 & 52.91 & 6.0989 \\
& Auxfree & 41.35 & 53.54 & 22.32 & 57.76 & 56.78 & 57.22 & 6.1140 \\
& Global-DO-loss & 41.51 & 54.96 & 22.75 & 58.35 & 57.14 & 58.27 & 6.0856 \\
\bottomrule
\end{tabular}
\caption{Validation perplexity and benchmark accuracy at two scales (100B and 200B tokens). At 30BA3B, we additionally evaluate Global-LBL for completeness; Global-LBL underperforms Auxfree on most benchmarks, particularly on knowledge-intensive tasks such as TriviaQA, indicating that load-balancing objectives can conflict with knowledge-intensive tasks at larger scales.}
\label{tab:main_results}
\end{table}
\section{Results and Analysis}
\label{sec:results}
\subsection{Regularization Coefficient Sensitivity}

Figure~\ref{fig:do_balancing} reveals a U-shaped relationship between the DO-loss coefficient and validation perplexity, with an optimal value at $\lambda = 10^{-5}$. This value is orders of magnitude smaller than the $10^{-3}$ to $10^{-4}$ range typical of load-balancing losses. The gap points to a structural difference: DO-loss simultaneously regularizes all expert pairs, and this high-dimensional pairwise constraint introduces substantially larger gradient variance than a scalar auxiliary loss over $n$ fractions. A much smaller coefficient is therefore required to prevent this variance from destabilizing training. Setting $\lambda$ too high causes the pairwise objective to overwhelm the cross-entropy signal, while setting it too low fails to encourage expert specialization and leads to expert collapse. A finer-grained sweep on a 2B/0.2B model (Appendix~\ref{appendix:lambda_sweep}) refines the optimum to $\lambda = 5{\times}10^{-6}$, roughly two orders of magnitude smaller than typical load-balancing loss coefficients.

\subsection{Downstream Benchmark Performance}

As shown in Table~\ref{tab:main_results}, Global-DO-loss leads on every metric at the 4.8BA0.5B scale. Orth-loss yields the worst performance, which stems directly from expert collapse preventing the model from utilizing its full capacity. Auxfree enforces load balance through per-expert bias terms that leave the main gradient undisturbed, avoiding interference from auxiliary losses and making it the strongest baseline at the 4.8BA0.5B scale. The performance gap between DO-loss and Global-DO-loss confirms that device-local routing statistics systematically underestimate cross-expert overlap. Due to compute constraints, we compare only against Auxfree and Global-LBL at the 30BA3B scale, where Global-DO-loss maintains a consistent advantage. Global-LBL underperforms Auxfree on most metrics at this scale, indicating that the gains of Global-DO-loss stem from orthogonalization rather than global statistics alone. Validation perplexity also improves monotonically with larger effective batch sizes for $G_{ij}$ statistics (see Appendix~\ref{appendix:batch_ablation} for the full sweep), consistent with the findings of \citet{qiu2025demons}.

\subsection{Ablation Study}

To isolate the contributions of the DO objective and global-batch statistics, we compare four variants on the 4.8BA0.5B model: local and global versions of both the load-balancing loss (Local-LBL) and DO-loss. As shown in Table~\ref{tab:main_results}, global statistics benefit both methods (Global-LBL $>$ Local-LBL, Global-DO-loss $>$ DO-loss). More importantly, the DO objective strictly dominates Local-LBL at both the local and global levels, with Global-DO-loss achieving the best performance across all seven metrics. This confirms that the pairwise orthogonalization objective, rather than the use of global statistics, is the primary driver of the observed improvements.

We further examine how the benefit of Global-DO-loss over Auxfree varies with expert granularity (see Appendix~\ref{appendix:expert_scaling} for full results). On a 2B/0.2B model, Global-DO-loss trails Auxfree at 32 experts but pulls ahead as granularity increases to 64 and 128 experts, consistent with the trend toward fine-grained architectures in modern LLMs.

\subsection{Expert Behavior: A Comparative Analysis}

\begin{figure}[htbp]
    \centering
    \begin{minipage}[c]{0.31\linewidth}
        \centering
        \includegraphics[width=\linewidth]{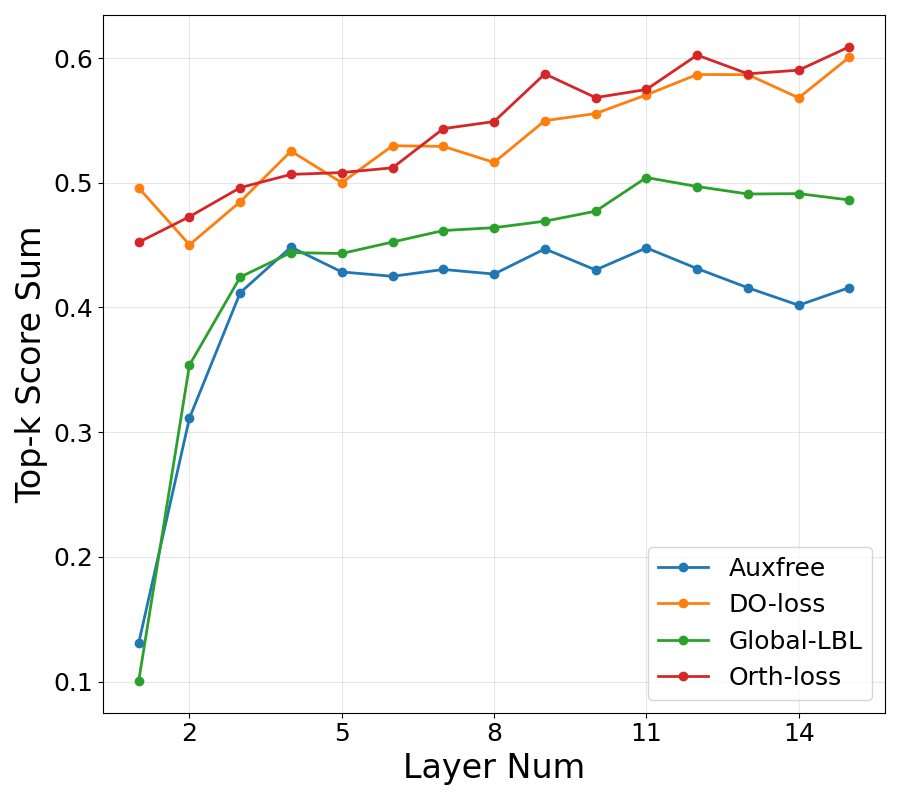}
        \centerline{\small(a) Top-$K$ score sum}
    \end{minipage}
    \hfill
    \begin{minipage}[c]{0.34\linewidth}
        \centering
        \includegraphics[width=0.48\linewidth]{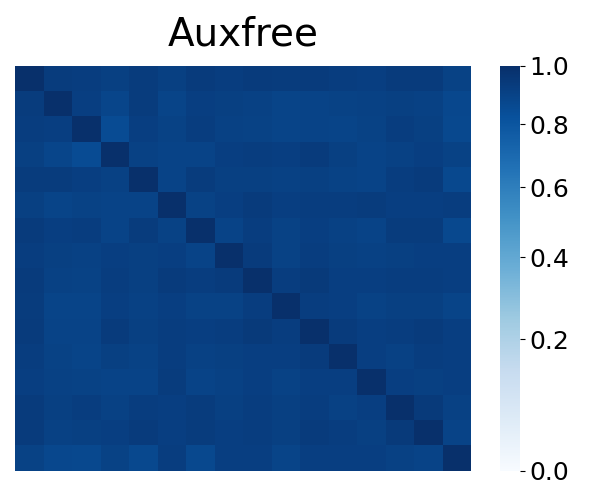}
        \includegraphics[width=0.48\linewidth]{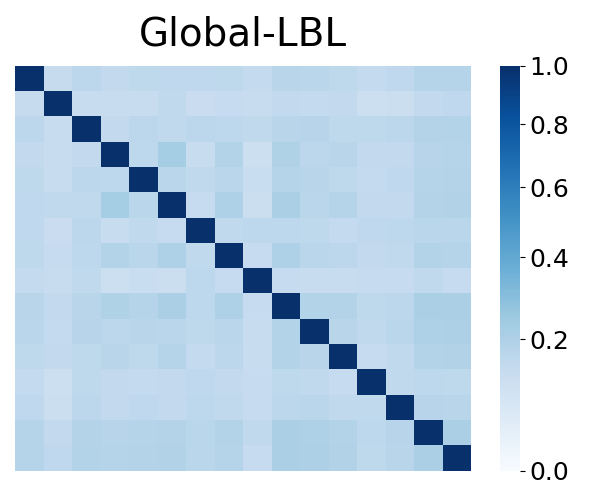}
        \includegraphics[width=0.48\linewidth]{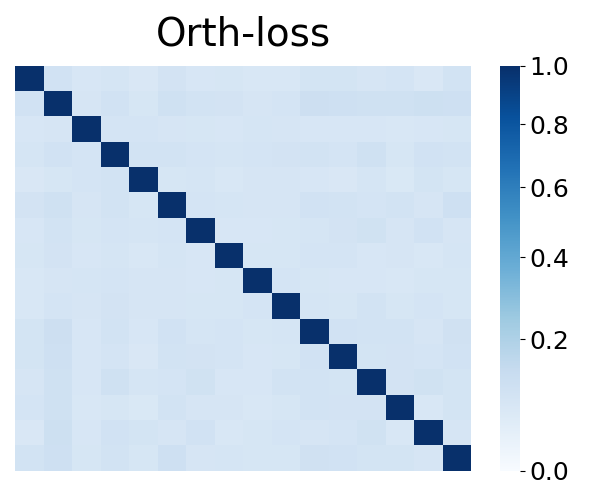}
        \includegraphics[width=0.48\linewidth]{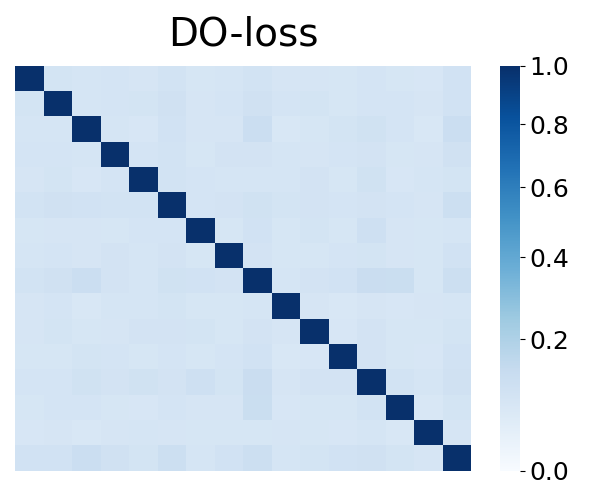}
        \centerline{\small(b) CKA similarity matrix}
    \end{minipage}
    \hfill
    \begin{minipage}[c]{0.33\linewidth}
        \centering
        \includegraphics[width=\linewidth]{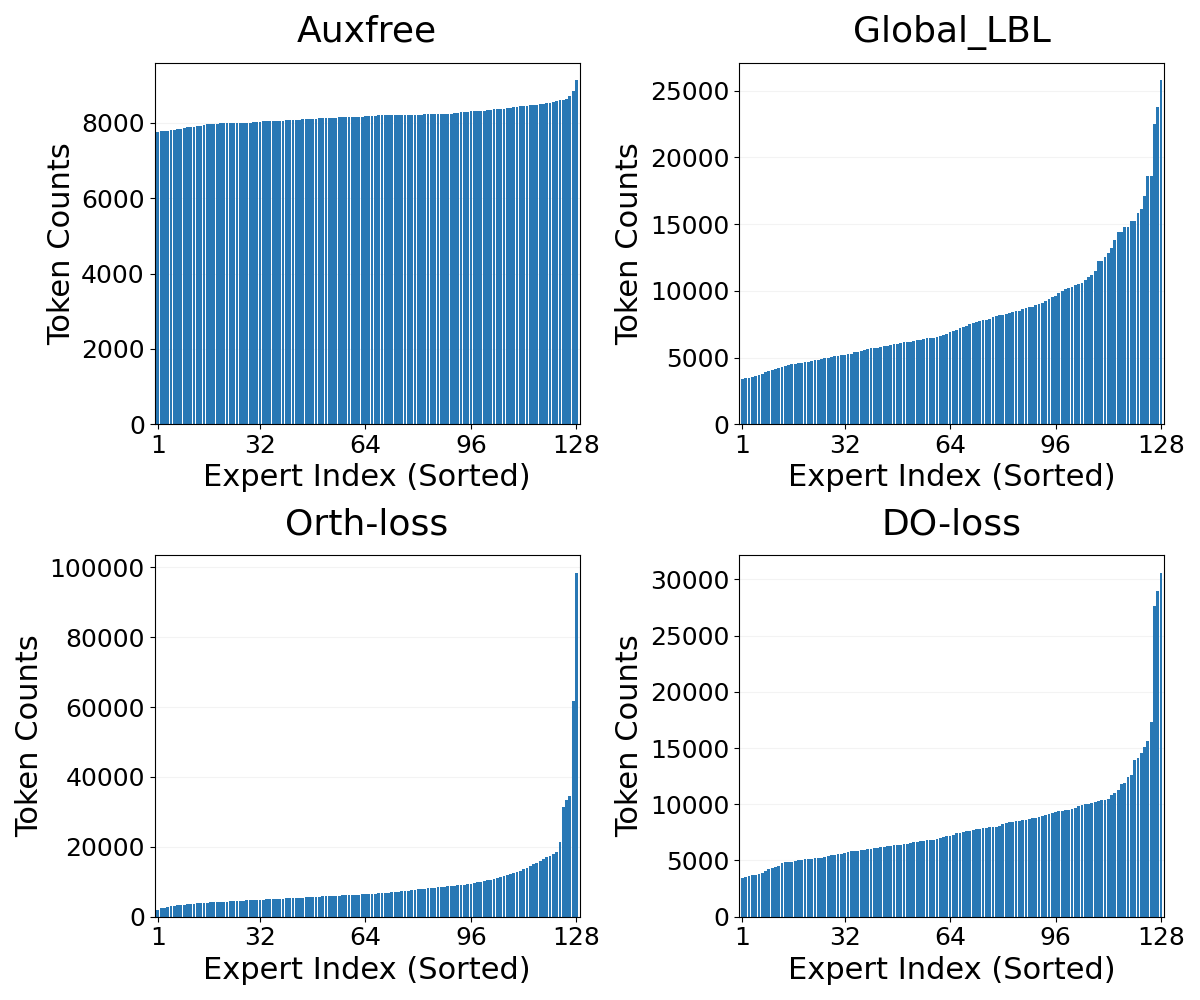}
        \centerline{\small(c) Load distributions}
    \end{minipage}
    \caption{Expert behavior in the trained 4.8BA0.5B model.
    \textbf{(a)} Top-$K$ routing score sum per token: higher values indicate the router assigns tokens with greater confidence, reflecting stronger expert specialization.
    \textbf{(b)} CKA similarity matrix (Appendix~\ref{appendix:cka}) at layer 4, sampled every 8th expert ($16{\times}16$ grids). CKA measures functional similarity between expert outputs on the same inputs; high off-diagonal values indicate representational redundancy, while a near-identity matrix indicates functional diversity.
    \textbf{(c)} Distribution of token counts per expert at layer 4.}
    \label{fig:routing_analysis}
\end{figure}

Figure~\ref{fig:routing_analysis} evaluates the methods across routing decisiveness (Panel a), inter-expert representational similarity (Panel b), and per-expert token load distribution (Panel c). None of the baselines satisfies all three criteria simultaneously, reflecting inherent design trade-offs.

Auxfree produces a nearly flat load histogram but yields the highest CKA off-diagonal values and lowest Top-$K$ sums, indicating that strict uniform token distribution suppresses the selective routing needed for expert differentiation. Global-LBL adopts global token statistics that allow some semantic clustering, reducing off-diagonal CKA and increasing Top-$K$ sums vs.~Auxfree, but its balancing objective still restricts full specialization.

Weight-space orthogonality promotes representational diversity without governing functional utilization. Orth-loss achieves the sparsest CKA and highest Top-$K$ sums, yet its token load exhibits a severely skewed exponential tail---parameter-space differentiation lacks constraints on token allocation, risking collapse despite near-ideal representational structures. DO-loss matches Orth-loss on CKA sparsity and Top-$K$ sums while maintaining a more controlled load distribution.

To complement the visual CKA comparison, we quantitatively measure CKA across all 128 expert pairs (Appendix~\ref{appendix:cka_quant}). The results confirm: Auxfree has the highest redundancy in shallow layers, Orth-loss achieves sparsest early-layer representations, and Global-DO-loss attains the best separation in deeper layers with smallest variance, consistent with its more decisive routing.

\subsection{REM: Load Balancing and System Efficiency}

\begin{figure}[htbp]
    \centering
    \includegraphics[width=0.85\linewidth]{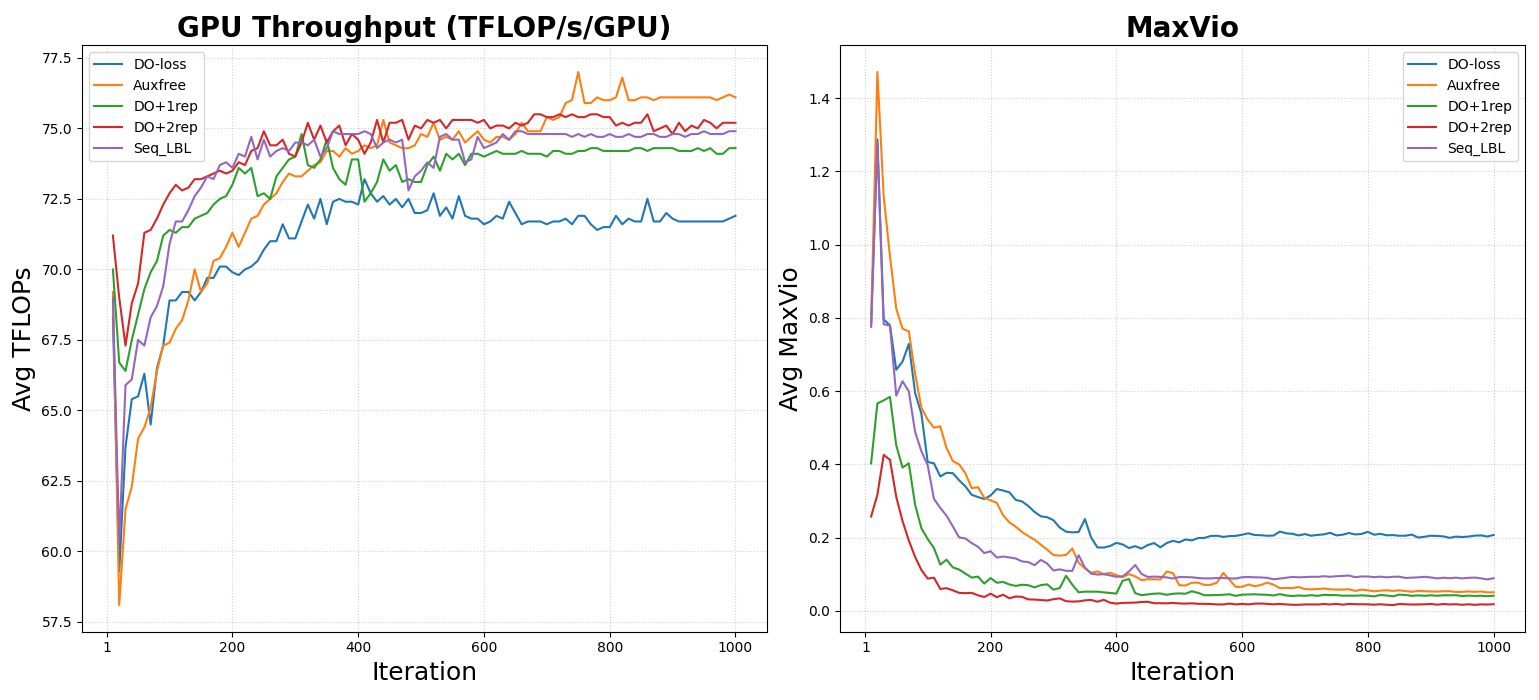}
    \caption{\textbf{Ablation of REM replica count.} GPU throughput and MaxVio over 1000 steps on H20 GPUs (EP$=8$, 4.8BA0.5B) with 0, 1, and 2 replica experts. A single replica recovers most of the throughput gap; two replicas drive MaxVio to near zero. MaxVio (Appendix~\ref{appendix:maxvio}) measures the fractional load excess of the busiest rank relative to the ideal average.}
    \label{fig:h20_efficiency}
\end{figure}

Figure~\ref{fig:h20_efficiency} shows Auxfree converging quickly to low MaxVio. Seq-LBL, which computes the auxiliary loss at the sequence level to enforce stricter per-sample balancing, settles at a slightly higher MaxVio. Unaugmented DO-loss remains persistently elevated---reflecting structural specialization, not slow convergence. REM resolves this: a single replica closes most of the throughput gap and drives MaxVio below Auxfree; two replicas achieve near-perfect balance. Both configurations maintain slightly lower throughput than Auxfree due to replica weight-copy overhead.

The memory cost of replication is modest and REM communication overhead is equally negligible: the AllGather for queue depths transmits only $|\mathcal{R}|$ integers per EP group. REM was enabled for all methods in throughput evaluations; for methods already achieving near-perfect load balance (Seq-LBL, Auxfree), REM provides only marginal additional gains.

We further compare REM against two representative expert-scheduling methods, EfficientMoE~\cite{zeng2025efficientmoe} and Pro-Prophet~\cite{wang2026pro}, on both H20 and H800 GPU clusters under two expert-parallelism configurations; REM consistently outperforms both baselines. Appendix~\ref{appendix:rem_overhead} provides full throughput results, overhead measurements, and a detailed analysis of the scheduling comparison.

\section{Conclusion}
\label{sec:conclusion}

MoE architectures scale model capacity by activating a token-specific expert subset, yet their effectiveness is limited by two intertwined challenges: expert collapse reduces the effective parameter count, while representation redundancy wastes the capacity that remains. Existing load balancing methods mitigate collapse and maintain system resource efficiency, but suppress specialization. Parameter orthogonalization, on the other hand, encourages functional diversity yet provides no mechanism to prevent collapse itself. Naively combining these independent solutions introduces conflicting gradients that destabilize training, revealing a deeper issue: both attempt to govern routing behavior through the training objective, coupling concerns that operate at fundamentally different levels.  
  
We argue that load balancing is not the only means to resolve collapse, and that system efficiency should not be enforced at the algorithmic level. To this end, we introduce DO-loss, which encourages specialization and mitigates collapse by penalizing routing trajectory overlap rather than weight similarity. For system efficiency, REM restores workload balance through adaptive replica placement and real-time token dispatching, entirely decoupled from the routing objective.

Experiments on both 4.8BA0.5B and 30BA3B MoE models show that this decoupled design consistently outperforms the evaluated routing algorithms on downstream tasks while maintaining comparable training throughput. This decoupling also makes the design adaptable to diverse hardware paradigms: on wafer-scale architectures where token routing and expert placement decouple differently from GPU expert parallelism, DO-loss alone can provide acceptable load balance, making REM optional. More broadly, our results suggest a design principle for future MoE systems: MoE effectiveness and system efficiency operate at fundamentally different abstraction levels requiring separate mechanisms. Routing objectives should focus solely on functional data partitioning, while load imbalance should be treated as an infrastructure problem addressed by system-level mechanisms such as replica scheduling and capacity-aware dispatching. Disentangling these concerns is key to ensuring that theoretical parameters reliably translate into actual model capacity.
\section*{Acknowledgements}
We thank Tencent for providing the computational resources and infrastructure that made this work possible. We also thank the anonymous reviewers for their insightful comments and constructive feedback, which helped improve the quality of this paper.

\bibliographystyle{colm2026_conference}
\bibliography{colm2026_conference} 

\newpage
\appendix

\section{DO-loss Normalization Coefficient}
\label{appendix:do_normalization}

We first clarify the definition of $p_{it}$. Let $s_{it}$ be the raw Sigmoid-gated routing score. Then $p_{it} = s_{it} / \sum_{i'} s_{i't}$ is the normalized routing probability, satisfying $\sum_i p_{it} = 1$. After Top-$K$ selection, only the probabilities of the $K$ chosen experts are retained and re-normalized. Consequently, $\bar{s}_{\text{all}} = \frac{1}{m}\sum_t\sum_i p_{it} = 1$ for every token, which is why this term naturally cancels in the derivation below.

We derive the normalization coefficient $\frac{1}{m^2k}$ in Eq.~\ref{eq:do_loss} by evaluating the scale of the unnormalized loss at the balanced equilibrium.

\textbf{Step 1: Sum of $G_{ij}$.}
$G_{ij} = \|F_i - F_j\|_1$, so the sum over all pairs follows from the zero-sum constraint (Eq.~\ref{eq:zero_sum}):
\begin{equation}
\sum_{i,j} G_{ij} = 2mk(n-k)
\end{equation}

\textbf{Step 2: Sum of $C_{ij}$.}
Expanding $C_{ij} = \sum_t p_{it}(f_{it} - f_{jt})$ and summing over all pairs:
\begin{align}
\sum_{i,j} C_{ij}
&= \sum_t \sum_i p_{it} \bigl( n f_{it} - \underbrace{\sum_j f_{jt}}_{=k} \bigr) \notag \\
&= n \sum_t \underbrace{\sum_i p_{it} f_{it}}_{\bar{s}_{K,t}} - km \cdot \underbrace{\tfrac{1}{m}\sum_t\sum_i p_{it}}_{\bar{s}_{\text{all}}}
\end{align}
where $\bar{s}_{K,t}$ is the sum of selected routing scores for token $t$ and $\bar{s}_{\text{all}}$ is the mean total score per token. Denoting $\bar{s}_K = \frac{1}{m}\sum_t \bar{s}_{K,t}$:
\begin{equation}
\sum_{i,j} C_{ij} = m(n\bar{s}_K - k)
\end{equation}

\textbf{Step 3: Scale at equilibrium.}
At the DO-loss minimum, all pairs share the same distance magnitude and gradient weight, so each $G_{ij}$ and $C_{ij}$ equals its mean:
\begin{equation}
\bar{G} = \frac{2mk(n-k)}{n^2}, \qquad \bar{C} = \frac{m(n\bar{s}_K - k)}{n^2}
\end{equation}
The unnormalized sum over all $n^2$ pairs is:
\begin{equation}
\mathcal{L}_{\text{unnorm}}
\approx n^2 \bar{G}\bar{C}
= \frac{2m^2 k(n-k)(n\bar{s}_K - k)}{n^2}
\end{equation}
Since $k \ll n$, we have $(n-k)\approx n$ and $(n\bar{s}_K-k)\approx n\bar{s}_K$, so:
\begin{equation}
\mathcal{L}_{\text{unnorm}} \approx 2m^2 k \bar{s}_K \sim \mathcal{O}(m^2 k)
\end{equation}
Dividing by $m^2k$ normalizes the loss to $\mathcal{O}(1)$ up to the constant $2\bar{s}_K$. Here $\bar{s}_K$ is the per-token mean of the Top-$K$ routing score sum. As specialization intensifies, routing concentrates, and the sum converges to $\bar{s}_K \approx 1$. Together with the factor of $2$ from pairwise symmetry, both are absorbed into the normalization. The resulting coefficient $\frac{1}{m^2k}$ makes DO-loss scale-free and directly comparable to the cross-entropy loss without further tuning.

\newpage
\section{Replica Expert Mechanism}
\label{appendix:rem_pseudocode}

REM uses two decoupled control loops operating at different time granularities, plus a training trajectory invariance mechanism that ensures mathematical equivalence with non-replicated execution.

\subsection{Global-Batch Replica Allocation}
\label{appendix:rem_allocation}

At each global-batch boundary, the controller greedily fills replica slots by repeatedly picking the least-loaded rank and placing the expert that maximizes the Benefit Score $S$ (Eq.~\ref{eq:benefit_score} in the main text), which represents the optimal number of tokens that the receiving replica $r_r$ can accept from $r_s$ to maximize load balance. Load estimates are updated after each placement until all $D$ slots per rank are filled or no beneficial placement remains.

\begin{algorithm}[htbp]
\caption{REM Global-Batch Replica Allocation}
\label{alg:rem_allocation}
\begin{tabbing}
\hspace{3em}\=\hspace{2em}\=\kill
\textbf{Require:} Historical rank loads $\{L_r\}$, expert token volumes $\{T_e\}$, per-rank capacity $D$ \\[2pt]
\textbf{Ensure:} Replica assignment $\mathcal{A} \leftarrow \emptyset$ \\[4pt]
$\tau \leftarrow \frac{1}{|\mathcal{R}|} \sum_{r} L_r$ \\
\textbf{repeat} \\
\> $r_r \leftarrow \arg\min_{\{r : |\mathcal{A}_r| < D\}} L_r$ \\
\> $(e^*, r_s^*) \leftarrow \arg\max_{e,\, r_s \neq r_r} S(e, r_s, r_r, \tau)$ \\
\> $\Delta \leftarrow S(e^*, r_s^*, r_r, \tau)$ \\
\> \textbf{if} $\Delta \leq 0$ \textbf{then break} \\
\> $\mathcal{A} \leftarrow \mathcal{A} \cup \{(e^*, r_s^*, r_r)\}$ \\
\> $L_{r_r} \mathrel{+}= \Delta$ \\
\> $L_{r_s^*} \mathrel{-}= \Delta$ \\
\> $T_{e^*} \mathrel{-}= \Delta$ \\
\textbf{until} all ranks full
\end{tabbing}
\end{algorithm}

\subsection{Micro-Batch Token Dispatching}
\label{appendix:rem_dispatch}

Within each micro-batch, a local dispatcher routes tokens independently of the global controller to prevent synchronization bottlenecks. After syncing queue depths via AllGather, the dispatcher repeatedly sorts all ranks by current load and traverses them from lightest to heaviest. For every rank, it assigns the highest-scoring replica tokens and updates the queue depth $Q_r$. Passes repeat $D$ times. The AllGather overhead is negligible compared to the AlltoAll dispatch, as it transmits only $|\mathcal{R}|$ integers per EP group.

\begin{algorithm}[htbp]
\caption{REM Micro-Batch Token Dispatching}
\label{alg:rem_dispatch}
\begin{tabbing}
\hspace{3em}\=\hspace{2em}\=\hspace{2em}\=\kill
\textbf{Require:} Replica assignment $\mathcal{A}$, real-time expert token counts $\{T_e\}$, per-rank capacity $D$ \\[2pt]
\textbf{Ensure:} Token-to-rank mapping $\mathcal{M} \leftarrow \emptyset$ \\[4pt]
$T_e \leftarrow \text{AllGather}(\{T_e\})$ \\
$Q_r \leftarrow \sum_{e \in r} T_e$ \\
$\tau \leftarrow \frac{1}{|\mathcal{R}|} \sum_{r} Q_r$ \\
\textbf{for} $d = 1$ \textbf{to} $D$ \textbf{do} \\
\> sort ranks by $Q_r$ ascending \\
\> \textbf{for} each rank $r$ in sorted order \textbf{do} \\
\> \> $(e^*, r_s^*) \leftarrow \arg\max_{(e,\,r_s)\,:\,(e,\,r_s,\,r)\in\mathcal{A}} S(e, r_s, r, \tau)$ \\
\> \> $\Delta \leftarrow S(e^*, r_s^*, r, \tau)$ \\
\> \> $\mathcal{M} \leftarrow \mathcal{M} \cup \{(r, e^*, \Delta)\}$ \\
\> \> $Q_{r} \mathrel{+}= \Delta$ \\
\> \> $Q_{r_s^*} \mathrel{-}= \Delta$ \\
\> \> $T_{e^*} \mathrel{-}= \Delta$ \\
\> \textbf{end for} \\
\textbf{end for}
\end{tabbing}
\end{algorithm}

\subsection{Training Trajectory Invariance}
\label{appendix:rem_invariance}

REM operates purely at the system level and preserves the mathematical training trajectory. The invariance relies on the following implementation design:

\textbf{Parameter synchronization.} Replica experts use independent memory allocations initialized to zero. Before each forward pass, replica parameters are overwritten with the current parameters of the original expert via asynchronous peer-to-peer communication. Fetching of early-layer parameters begins before the forward pass, and subsequent layers are fetched during the computation of prior layers, overlapping communication with computation.

\textbf{Gradient accumulation.} Each micro-batch computes gradients independently on all replicas. After each global step, the accumulated gradients from all replicas are sent back to the original expert rank and summed. Because replicas hold a frozen copy of the original parameters at the time of the forward pass, the gradients computed remotely are numerically identical to those that would have been computed locally.

\textbf{Optimizer state.} Optimizer states (e.g., Adam moments $\exp\_\text{avg}$ and $\exp\_\text{avg}\_\text{sq}$) reside exclusively on the original expert rank. Standard DDP all-reduce is applied only to the accumulated gradient on the original expert. A 10,000-step precision test confirmed gradient alignment to three decimal places.

\textbf{Equivalence conditions.} The training trajectory equivalence holds under three assumptions: (a)~replica parameters are fully synchronized with the original expert before each forward pass, (b)~gradient accumulation across replicas uses summation, and (c)~optimizer updates occur only on the original expert rank.

\newpage
\section{Gradient Analysis of DO-loss}
\label{appendix:gradient_analysis}

\subsection{Zero-Sum Constraint and Variance Minimization}

Under Top-$K$ routing with $n$ experts and $K$ selections per token over a batch of $m$ tokens, the total number of bits set across all expert load signatures is exactly $mK$. The sum of all pairwise Hamming distances is therefore constant:
\begin{equation}
\sum_{i=1}^n \sum_{j=1}^n \| F_i - F_j \|_1 = 2mK(n-K)
\end{equation}
Minimizing the sum of squared pairwise distances $\sum_{i\neq j} \|F_i - F_j\|_1^2$ under this fixed-sum constraint is equivalent to minimizing the variance of the pairwise distances. The global minimum is achieved when all pairwise distances are equal to $d^* = \frac{2mK(n-K)}{n^2}$.

\subsection{Anti-Collapse Gradient}

When expert $i$ is idle ($F_i = \mathbf{0}$), its distance to any active expert $j$ ($F_j \neq \mathbf{0}$) is $\|F_i - F_j\|_1 = \|F_j\|_1 > 0$. Under the zero-sum constraint, these nonzero distances increase the variance of the pairwise distance distribution. The gradient of the DO-loss with respect to the routing probability $p_{it}$ for an idle expert $i$ is:
\begin{equation}
\nabla_{p_{it}} \mathcal{L}_{\text{DO}} \propto -\sum_{j: F_j \neq \mathbf{0}} \|F_i - F_j\|_1 \cdot f_{jt}
\end{equation}
The negative sign indicates that the loss pushes $p_{it}$ to increase for tokens that are currently assigned to active experts. As the router raises these probabilities, the idle expert begins to receive tokens, which reduces the distances between it and the previously active experts and thereby lowers the overall variance. This negative feedback loop makes expert collapse self-correcting without requiring an explicit load-balancing term.

\subsection{Equidistance and Optimal Specialization}

We now formalize the relationship between pairwise equidistance and expert specialization. Let $c_{ij} = \sum_t f_{it} f_{jt}$ denote the co-occurrence count of experts $i$ and $j$ (the number of tokens they both select). The Hamming distance can be expressed as:
\begin{equation}
\|F_i - F_j\|_1 = \|F_i\|_1 + \|F_j\|_1 - 2c_{ij}
\end{equation}

When all pairwise distances are equal ($\|F_i - F_j\|_1 = d^*$ for all $i \neq j$), summing this equality over all $j \neq i$ for a fixed $i$ yields $\|F_i\|_1 = \frac{mK}{n}$ for every expert $i$. That is, equidistance implies uniform per-expert token counts. Substituting into the Hamming distance formula gives $c_{ij} = \frac{mK(K-1)}{n(n-1)}$ for all $i \neq j$, which is the minimum possible co-occurrence under the constraints of Top-$K$ routing. This means that under the equidistance condition, every pair of experts shares the smallest possible number of tokens.

Since the router naturally converges to assign semantically similar tokens to the same expert subsets, this minimal and uniform co-occurrence ensures that distinct semantic patterns are processed by maximally disjoint combinations of experts. The result is a functionally diversified expert population in which each expert develops a specialized role.

\newpage
\section{Centered Kernel Alignment (CKA)}
\label{appendix:cka}

\textbf{Feature extraction.}
Each expert is a SwiGLU FFN~\cite{shazeer2020glu} with input projections $W_1^{(i)}, W_3^{(i)} \in \mathbb{R}^{d \times d_e}$ and output projection $W_2^{(i)} \in \mathbb{R}^{d_e \times d}$. For input $\mathbf{x} \in \mathbb{R}^d$:
\begin{equation}
\text{Expert}_i(\mathbf{x}) = \bigl(\sigma(\mathbf{x} W_1^{(i)}) \odot \mathbf{x} W_3^{(i)}\bigr) W_2^{(i)}
\end{equation}
where $\sigma(\cdot)$ is SiLU. We use the intermediate gated representation $\sigma(\mathbf{x}W_1^{(i)}) \odot \mathbf{x}W_3^{(i)}$ as the feature vector: it combines gating magnitude with the value branch and therefore captures each expert's functional selectivity more faithfully than either branch alone or the final output.

A fixed probe set of $m$ tokens is passed through every expert independently (bypassing the router). For expert $i$:
\begin{equation}
\mathbf{A}_i = \bigl[\sigma(\mathbf{x}_1 W_1^{(i)}) \odot \mathbf{x}_1 W_3^{(i)};\cdots;\sigma(\mathbf{x}_m W_1^{(i)}) \odot \mathbf{x}_m W_3^{(i)}\bigr]^\top \in \mathbb{R}^{m \times d_e}
\end{equation}

\textbf{Linear CKA.}
Linear CKA~\citep{kornblith2019similarity} compares the centered Gram structures of two feature matrices. First, the feature matrix $\mathbf{A}_i$ is centered along the token axis:
\begin{equation}
\tilde{\mathbf{A}}_i = \left(\mathbf{I}_m - \tfrac{1}{m}\mathbf{1}_m\mathbf{1}_m^\top\right) \mathbf{A}_i
\end{equation}
To better illustrate the distinction between the Global-LBL approach and orthogonalization methods, we compute the square root of the standard metric. The final evaluation value is defined as:
\begin{equation}
\text{CKA}(\mathbf{A}_i, \mathbf{A}_j)
= \sqrt{ \frac{\|\tilde{\mathbf{A}}_i^\top \tilde{\mathbf{A}}_j\|_F^2}
       {\|\tilde{\mathbf{A}}_i^\top \tilde{\mathbf{A}}_i\|_F \|\tilde{\mathbf{A}}_j^\top \tilde{\mathbf{A}}_j\|_F} }
\end{equation}
Within the square root, the numerator represents the Hilbert--Schmidt inner product of the centered Gram matrices, while the denominator provides normalization to ensure invariance to isotropic scaling. A resulting value of $0$ indicates that the two experts produce linearly independent representations, whereas a value of $1$ signifies that the representations are identical up to an invertible linear transformation. Consequently, high off-diagonal values indicate that distinct experts extract highly correlated features for the same tokens, which serves as a direct symptom of representation redundancy.

\textbf{Quantitative results.}
\label{appendix:cka_quant}
To complement the qualitative CKA heatmaps in the main text, Table~\ref{tab:cka_quant} reports the mean and standard deviation of CKA similarity across all 128 expert pairs at Layers~4 and~8. Adopting 0.1 as a threshold for excessive similarity, Auxfree exhibits the highest redundancy, particularly in the shallower layer where its mean CKA substantially exceeds the threshold. Orth-loss achieves the sparsest representations in Layer~4, while Global-DO-loss attains the lowest mean and smallest variance in the deeper Layer~8, consistent with its more decisive and diversified routing.

\begin{table}[h]
\centering
\small
\begin{tabular}{@{}l c c@{}}
\toprule
Method & Layer~4 CKA mean (std) & Layer~8 CKA mean (std) \\
\midrule
Auxfree & 0.263 (0.031) & 0.115 (0.023) \\
Global-LBL & 0.089 (0.016) & 0.077 (0.008) \\
Orth-loss & \textbf{0.084} (0.010) & 0.084 (0.011) \\
Global-DO-loss & 0.086 (0.010) & \textbf{0.071} (0.007) \\
\bottomrule
\end{tabular}
\caption{Mean and standard deviation of CKA cosine similarity across all 128 expert pairs. Lower values indicate less representational redundancy. Best results in \textbf{bold}.}
\label{tab:cka_quant}
\end{table}

\newpage
\section{MaxVio Metric}
\label{appendix:maxvio}

We adopt \textbf{Maximum Violation (MaxVio)} to evaluate the degree of outlier behavior at the rank level. Specifically, each rank $r$ within an EP group has a total load $L_r$, calculated as the sum of tokens across all experts residing on that rank. MaxVio measures the extent to which the load of the busiest rank deviates from the ideal uniform load $\tau$:
\begin{equation}
\text{MaxVio} = \max_{r \in \mathcal{R}} \left| \frac{L_r}{\tau} - 1 \right|, \qquad \tau = \frac{1}{|\mathcal{R}|} \sum_{r \in \mathcal{R}} L_r
\end{equation}
The reported value represents the average across all MoE layers and micro-batches during a single training step. This metric directly captures the straggler effect, as any rank processing more tokens than $\tau$ stalls the entire EP group. For instance, a MaxVio value of $0.1$ indicates that the most heavily loaded rank processes $10\%$ more tokens than the ideal distribution on average.

\begin{figure}[h]
    \centering
    \includegraphics[width=0.65\linewidth]{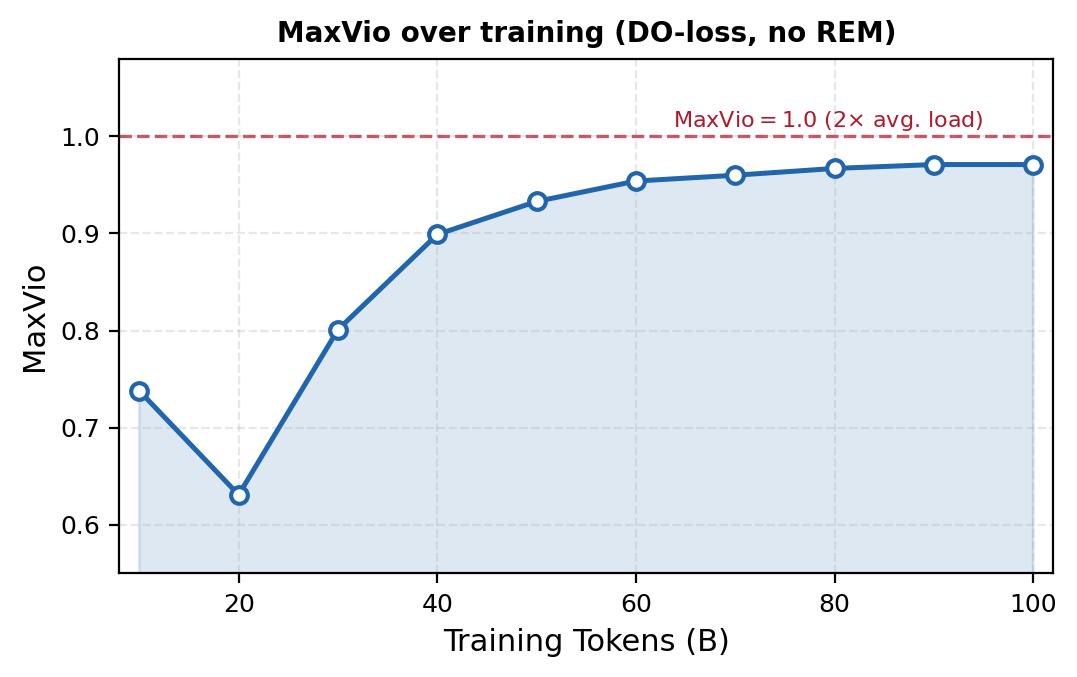}
    \caption{\textbf{Ablation of load imbalance without REM.} MaxVio over the course of 100B-token training for DO-loss alone (no replica scheduling). MaxVio stabilizes below 1.0, indicating that even without REM the busiest rank processes at most twice the average load, though it remains substantially higher than with replica scheduling enabled.}
    \label{fig:maxvio_training}
\end{figure}

\newpage
\setcounter{section}{5}
\section{Supplementary Experiments}
\label{appendix:supplementary}

This section collects auxiliary experiments and implementation details that support claims made in the main text.

\subsection{Fine-grained $\lambda$ Sweep}
\label{appendix:lambda_sweep}

\begin{figure}[h]
    \centering
    \includegraphics[width=0.55\linewidth]{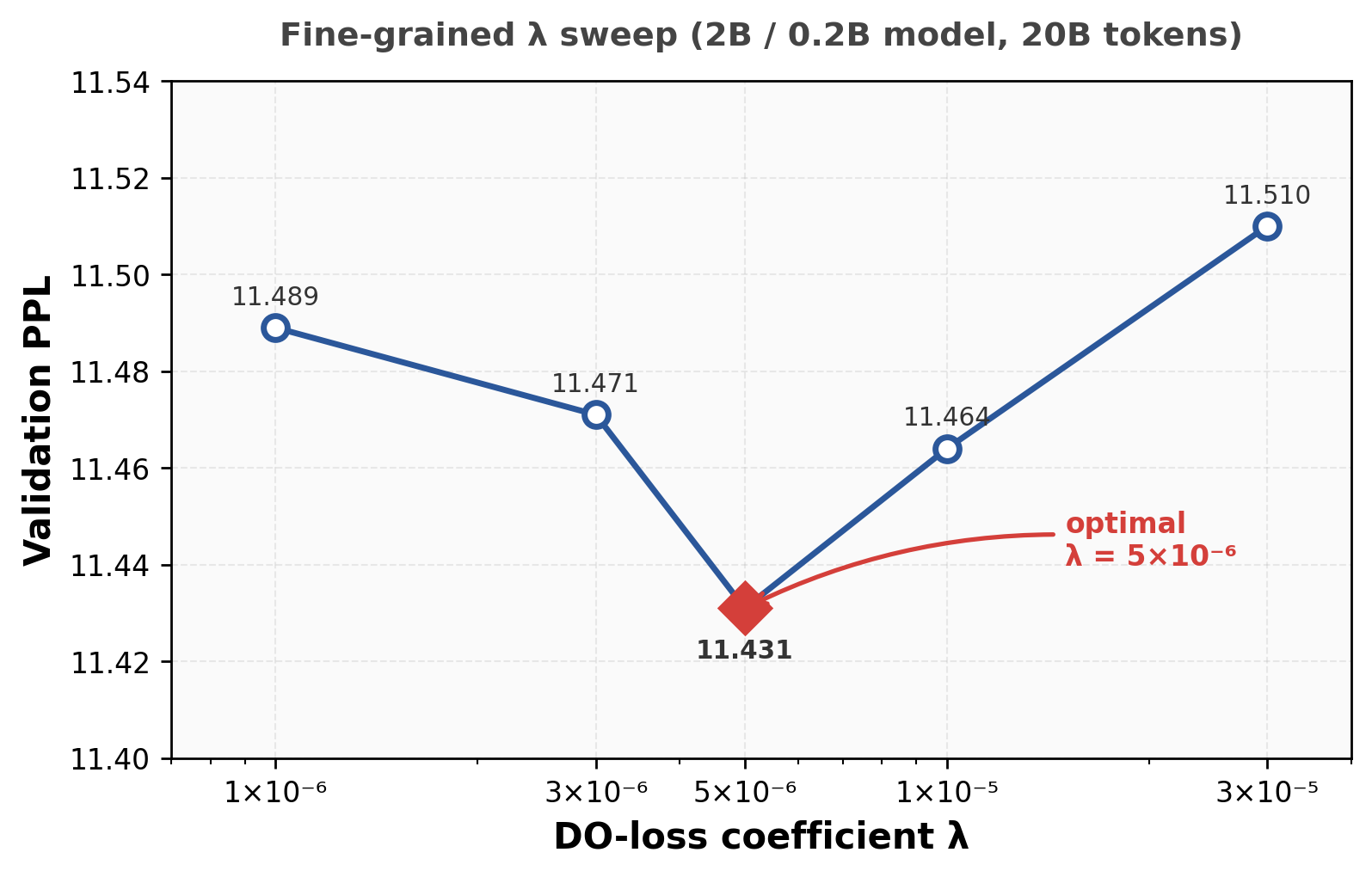}
    \caption{\textbf{Ablation of DO-loss coefficient $\lambda$ at finer resolution.} Sweep on a 2B/0.2B model (20B tokens). The optimum $\lambda{=}5{\times}10^{-6}$ is roughly two orders of magnitude smaller than typical load-balancing loss coefficients.}
    \label{fig:lambda_sweep}
\end{figure}

To pinpoint the optimal DO-loss coefficient at finer resolution, we sweep $\lambda \in \{1\times10^{-6}, 3\times10^{-6}, 5\times10^{-6}, 1\times10^{-5}, 3\times10^{-5}\}$ on a 2B/0.2B model trained for 20B tokens. Figure~\ref{fig:lambda_sweep} shows that the optimum lies at $\lambda = 5\times10^{-6}$, roughly two orders of magnitude smaller than typical load-balancing loss coefficients, confirming the trend observed on the main 4.8B model.

\subsection{DO-loss Batch Size Ablation}
\label{appendix:batch_ablation}

Table~\ref{tab:batch_ablation} ablates the effective batch size for computing the global distance matrix $G_{ij}$ on the 4.8BA0.5B model, keeping the micro-batch size fixed at 4 and synchronizing across 8 GPUs. Validation perplexity improves monotonically as the statistical scope widens from a single-device micro-batch to the full global batch, but with diminishing marginal returns, consistent with the batch-size effects documented by \citet{qiu2025demons} for load-balancing losses.

\begin{table}[h]
\centering
\small
\setlength{\tabcolsep}{4pt}
\begin{tabular}{@{}l c c c c c@{}}
\toprule
Configuration & Micro-batch & GPUs & grad\_accum & Eff.\ Batch ($G_{ij}$) & Val.\ PPL \\
\midrule
DO-loss (local) & 4 & -- & -- & 4 (no sync) & 8.4518 \\
DO-32 & 4 & 8 & 1 & 32 & 8.4475 \\
DO-64 & 4 & 8 & 2 & 64 & 8.4458 \\
DO-128 & 4 & 8 & 4 & 128 & 8.4446 \\
Global-DO-loss & 4 & 8 & 16 & 512 & 8.4433 \\
\bottomrule
\end{tabular}
\caption{Effect of the effective batch size for $G_{ij}$ statistics on validation perplexity (4.8BA0.5B, 100B tokens).}
\label{tab:batch_ablation}
\end{table}

\subsection{Expert Count Scaling}
\label{appendix:expert_scaling}

\begin{figure}[h]
    \centering
    \includegraphics[width=0.58\linewidth]{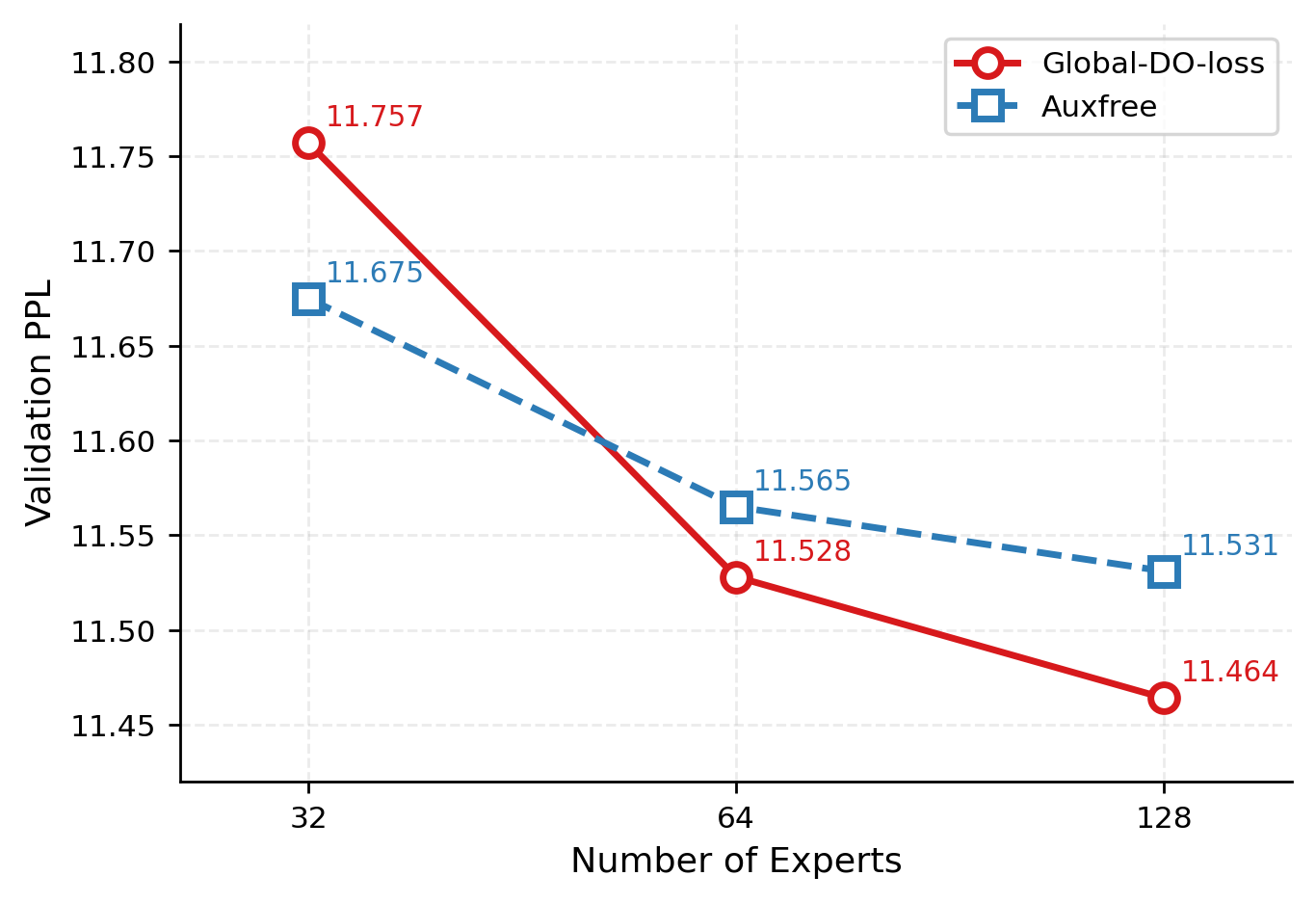}
    \caption{\textbf{Ablation of expert granularity.} Validation perplexity vs.~expert count (2B/0.2B, 20B tokens). Global-DO-loss trails Auxfree at 32 experts but surpasses it as granularity increases to 64 and 128, demonstrating that DO-loss benefits more from finer expert specialization.}
    \label{fig:expert_scaling}
\end{figure}

Figure~\ref{fig:expert_scaling} compares Global-DO-loss and Auxfree on a 2B/0.2B model as the number of experts increases from 32 to 128. At 32 experts, Auxfree holds a slight edge; however, as expert granularity increases, Global-DO-loss pulls ahead and the gap widens. This demonstrates that the benefit of distributional orthogonalization grows with the number of experts---precisely the regime where expert collapse and representational redundancy become more severe. The result aligns with the trend toward fine-grained expert architectures in modern LLMs.

\subsection{REM Memory and Throughput Overhead}
\label{appendix:rem_overhead}

Table~\ref{tab:scheduling} reports the steady-state throughput of each scheduling method on both H20 and H800 GPU clusters. All methods use two replica experts per rank on top of the DO-loss-trained 4.8BA0.5B model. REM consistently outperforms EfficientMoE and Pro-Prophet across all configurations and approaches the Auxfree baseline.

\textbf{Throughput recovery.} EP8 yields higher throughput than EP16 across all methods due to the absence of inter-node all-to-all communication. The DO-only baseline incurs a throughput penalty of roughly 10\% relative to Auxfree under EP8, and a larger drop under EP16 due to exacerbated inter-node imbalance. REM recovers the vast majority of this gap across all configurations, bringing throughput to within a few percent of the Auxfree baseline. Under EP16, REM nearly matches Auxfree because replica experts inherently reduce costly inter-node communication by routing tokens to local replicas.

\begin{table}[h]
\centering
\small
\setlength{\tabcolsep}{3pt}
\begin{tabular}{@{}l c c c c@{} }
\toprule
& \multicolumn{2}{c}{H20} & \multicolumn{2}{c}{H800} \\
\cmidrule(lr){2-3} \cmidrule(lr){4-5}
Method & DP2 EP8 & DP1 EP16 & DP2 EP8 & DP1 EP16 \\
\midrule
Auxfree & 81.2 & 70.1 & 161.3 & 101.8 \\
DO-only & 75.3 & 62.8 & 152.7 & 93.1 \\
DO+REM & 79.5 & 69.7 & 159.6 & 101.3 \\
DO+EfficientMoE & 77.8 & 67.5 & 155.2 & 95.7 \\
DO+Prophet & 76.8 & 65.9 & 152.3 & 92.2 \\
\bottomrule
\end{tabular}
\caption{\textbf{Ablation of expert scheduling method.} Steady-state throughput (TFLOPs) averaged over 1,000 steps. All methods use two replica experts per rank on the DO-loss-trained 4.8BA0.5B model.}
\label{tab:scheduling}
\end{table}

\textbf{Scheduling comparison.} EfficientMoE lags behind REM because it operates only at the global-batch level without real-time scheduling, leaving transient intra-step imbalances unresolved. Pro-Prophet introduces micro-batch-level replica replication whose communication overhead, under communication-bound conditions, can offset its scheduling benefits. In contrast, REM's two-tiered design---global replica placement plus lightweight micro-batch dispatching---avoids these pitfalls.

\textbf{Memory overhead.} Peak GPU memory increases modestly (single-digit percentage) with one or two replica experts, as replicas add only static parameter copies without extra activation memory. This overhead is negligible at the scales where MoE models are typically deployed.

\textbf{Communication overhead.} The AllGather for queue depths transmits only $|\mathcal{R}|$ integers per EP group, which is dwarfed by the AlltoAll dispatch that moves full hidden states.

\end{document}